\documentclass[pdflatex,sn-mathphys-num]{sn-jnl}

\usepackage{graphicx}%
\usepackage{multirow}%
\usepackage{amsmath,amssymb,amsfonts}%
\usepackage{amsthm}%
\usepackage{mathrsfs}%
\usepackage[title]{appendix}%
\usepackage{xcolor}%
\usepackage{textcomp}%
\usepackage{manyfoot}%
\usepackage{booktabs}%
\usepackage{algorithm}%
\usepackage{algorithmicx}%
\usepackage{algpseudocode}%
\usepackage{listings}%

\theoremstyle{thmstyleone}%
\theoremstyle{thmstyletwo}%

\theoremstyle{thmstylethree}%

\begin{document}

\title{Lessons from Neuroscience for AI: How integrating Actions, Compositional Structure and Episodic Memory could enable Safe, Interpretable and Human-Like AI}


\author*[1]{\fnm{Rajesh P.N.} \sur{Rao}}\email{rao@cs.washington.edu}

\author[1]{\fnm{Vishwas} \sur{Sathish}}\email{vsathish@cs.washington.edu}

\author[1]{\fnm{Linxing Preston} \sur{Jiang}}\email{prestonj@cs.washington.edu}

\author[1]{\fnm{Matthew} \sur{Bryan}}\email{mmattb@cs.washington.edu}

\author[1]{\fnm{Prashant} \sur{Rangarajan}}\email{prashr@cs.washington.edu}

\affil*[1]{\orgdiv{Paul G. Allen School of Computer Science}, \orgname{University of Washington}, \orgaddress{
\city{Seattle}, \postcode{98195}, \state{WA}, \country{USA}}}


\abstract{The phenomenal advances in large language models (LLMs) and other foundation models over the past few years have been based on optimizing large-scale transformer models on the surprisingly simple objective of minimizing next-token prediction loss, a form of predictive coding that is also the backbone of an increasingly popular model of brain function in neuroscience and cognitive science. However, current foundation models ignore three other important components of state-of-the-art predictive coding models: tight integration of actions with generative models, hierarchical compositional structure, and episodic memory. We propose that to achieve safe, interpretable, energy-efficient, and human-like AI, foundation models should integrate actions, at multiple scales of abstraction, with a compositional generative architecture and episodic memory. We present recent evidence from neuroscience and cognitive science on the importance of each of these components. We describe how the addition of these missing components to foundation models could help address some of their current deficiencies: hallucinations and superficial understanding of concepts due to lack of grounding, a missing sense of agency/responsibility due to lack of control, threats to safety and trustworthiness due to lack of interpretability, and energy inefficiency. We compare our proposal to current trends, such as adding chain-of-thought (CoT) reasoning  and retrieval-augmented generation (RAG) to foundation models, and discuss new ways of augmenting these models with  brain-inspired components. We conclude by arguing that a rekindling of the historically fruitful exchange of ideas between brain science and AI will help pave the way towards safe and interpretable human-centered AI. }




\maketitle

\section{Introduction}\label{sec1}

Predictive coding (or more generally, ``predictive processing'') \cite{Rao_Ballard_1999, Jiang_Rao_2022, Keller_Mrsic-Flogel_2018} has emerged in recent years as an important framework for understanding brain function. Theories based on predictive coding, which include active inference and the free energy principle \cite{friston2009fep, friston2009reinforcement,friston2010fep, Bogacz_2017}, postulate that the brain's goal is to learn an internal model of the world and use that model to predict upcoming inputs. The internal model is updated, during both inference and learning, by minimizing prediction errors. 

Over the past decade, while neuroscientists were engaged in searching for predictive signals and prediction errors in the brain \cite{Schneider_Sundararajan_Mooney_2018,Jordan_Keller_2020,Furutachi_Franklin_Aldea_Mrsic-Flogel_Hofer_2024}, researchers in AI independently discovered that a simple objective based on autoregressively predicting the next input token and minimizing the resulting prediction errors through backpropagation allows self-supervised learning of {\em foundation models} \cite{Bommasani2021risks}, such as large language models (LLMs), using transformer neural networks and internet-scale data \cite{vaswani2017transformers, Radford_Narasimhan_Salimans_Sutskever_2018,Radford_Wu_Child_Luan_Amodei_Sutskever_2019}.\footnote{We focus here mostly on transformer-based architectures trained with autoregressive loss rather than diffusion-based foundation models \cite{Song_etal_2020, Rombach_Blattmann_Lorenz_Esser_Ommer_2022, Li_Thickstun_Gulrajani_Liang_Hashimoto_2022}.} Transformers employ a scaled-dot-product attention mechanism \cite{vaswani2017transformers}, itself inspired by the concept of attention in human cognition, for learning rich and versatile embeddings of data. They have been applied to a variety of modalities - text, images, audio, video, timeseries, etc. \cite{vaswani2017transformers, dosovitskiy2021vit, baevski2020wave2vec, bertasius2021vid, zhou2021informer}. LLMs have also been shown to exhibit internal representations that are human-like in sentence processing \cite{Kuribayashi-2025} and other cognitive tasks \cite{Hu-et-al-2025} (see also \cite{Michaelov-2024}).

More recently, ``chain-of-thought'' (CoT) and test-time scaling have emerged as powerful techniques for enhancing LLM reasoning capabilities \cite{Wei_Wang_Schuurmans_Bosma_Ichter_Xia_Chi_Le_Zhou_2022, OpenAI_2024, Brown_Juravsky_Ehrlich_Clark_Le_Ré_Mirhoseini_2024, Snell_Lee_Xu_Kumar_2024, Muennighoff_Yang_Shi_Li_Fei-Fei_Hajishirzi_Zettlemoyer_Liang_Candès_Hashimoto_2025}. CoT allows language models to ``think'' step-by-step, improving the quality and accuracy of predicted outputs \cite{Wei_Wang_Schuurmans_Bosma_Ichter_Xia_Chi_Le_Zhou_2022}. More generally, scaling test-time compute can allow these models to optimize a trade-off between computing cost and accuracy \cite{Snell_Lee_Xu_Kumar_2024}. Additionally, fine-tuning techniques based on reinforcement learning (RL) can be used to reward a pre-trained model when output sequences lead to the correct answer, eliciting reasoning-like capabilities and improved performance \cite{OpenAI_2024, shao2024deepseekmathpushinglimitsmathematical, deepseekai2025deepseekr1incentivizingreasoningcapability}. 

However, LLMs and other foundation models currently ignore three important components that have proved useful in state-of-the-art predictive coding models of the brain: tight integration of actions with generative models, hierarchical compositional structure, and episodic memory. Here we argue that incorporating these components into foundation models could lead to safe, interpretable, energy-efficient, and human-like AI. 

In Section~\ref{sec:predcoding}, we provide background for the rest of the paper by presenting the predictive coding framework, highlighting some similarities and differences with transformer-based foundation models. In Sections~\ref{sec:actions}-\ref{sec:memory}, we present recent evidence from neuroscience and cognitive science on the importance of each of the above three components. We argue that the addition of each of these missing components to foundation models could help address their deficiencies such as hallucinations and superficial understanding of concepts due to lack of grounding \cite{West-et-al-2024paradox,Pezzulo2024meaning}, susceptibility to bias \cite{Navigli2023bias} and a missing sense of agency/responsibility due to lack of control \cite{Pezzulo2024meaning}, threats to safety and trustworthiness due to lack of interpretability \cite{Bommasani2021risks}, and poor energy efficiency \cite{crownhart2024energyuse}. We describe  current trends in the design of foundation models, such as adding chain-of-thought (CoT) reasoning  and retrieval-augmented generation (RAG), and suggest new ways of augmenting these models with brain-inspired components. 
We conclude in Section~\ref{sec:discussion} by calling for a rekindling of the historically productive synergy between brain science and AI to guide the development of safe, interpretable, energy-efficient human-like AI.

\section{Predictive Coding Theories of Brain Function} \label{sec:predcoding}
\begin{figure}
\centering
\includegraphics[width=0.9\linewidth]{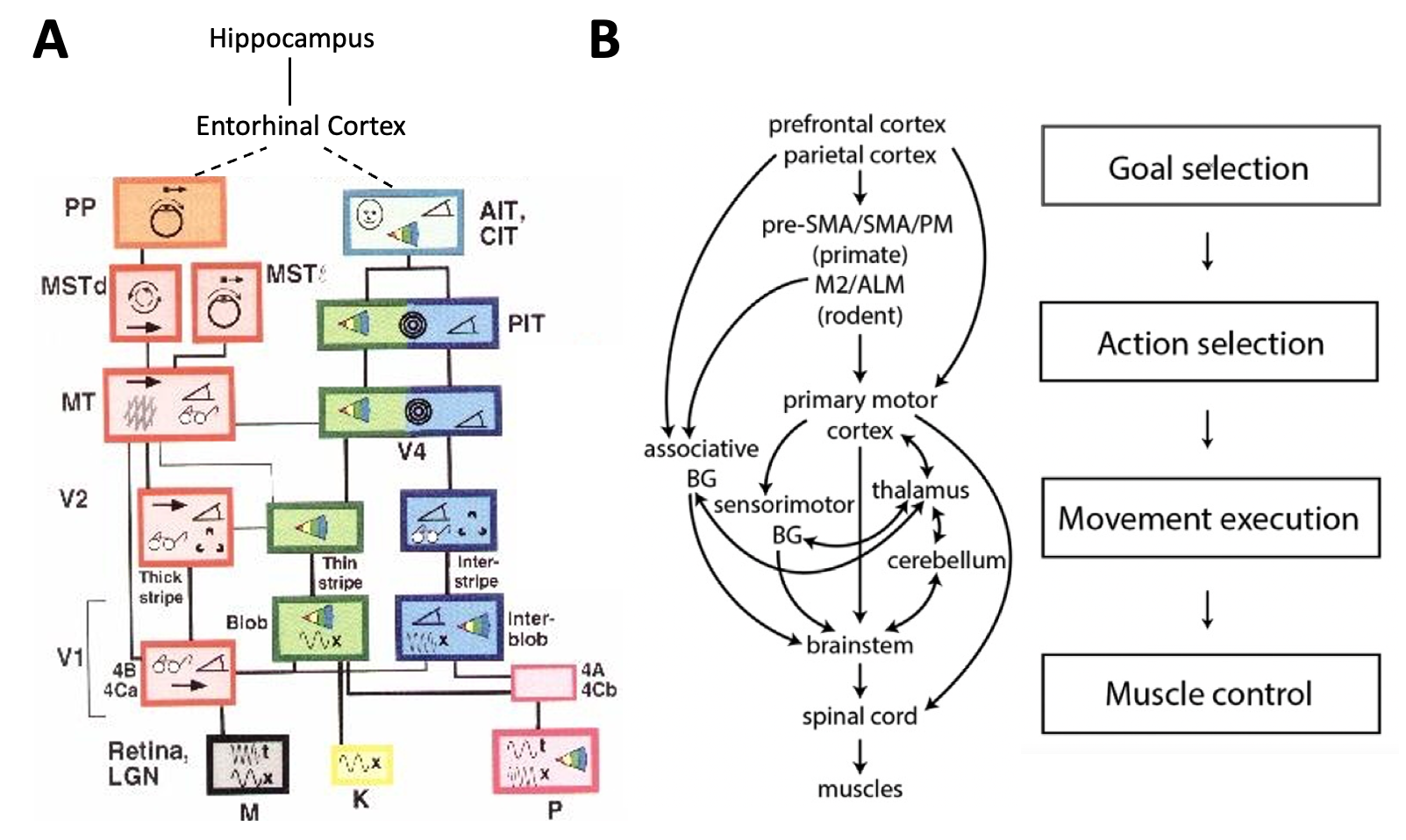}
\caption{
\textbf{The Hierarchical Brain}. (A) Hierarchy in the visual system. (Left) Bidirectionally connected areas of the cortex (in the middle) receiving visual information from the retina (bottom) and storing/receiving episodic context from the hippocampus (at the top). (Right) Simplified version of the cortical hierarchy of visual areas (V1, V2, V4/MT, PIT/MST, etc.). The areas encode increasingly  abstract features as we move to the top. (Adapted from \cite{Felleman_VanEssen_1991,VanEssen_1994}) (B) Hierarchy in the motor system. (Left) Hierarchical organization of brain regions implicated in movement. (Right) Abstraction level of processing in these regions. (Adapted from \cite{Olveczky-hier2022}) 
}
\label{fig:hier-brain}
\end{figure}

Predictive coding theories assume that the brain maintains an internal model of the world, and performs both inference and learning by minimizing prediction errors. A primary motivation for adopting such a view is the observation that the architecture of the brain is hierarchical (Figures~\ref{fig:hier-brain}A and B) with ubiquitous feedback connections \cite{zeki1988functional, Felleman_VanEssen_1991, angelucci2002circuits}: every area in the neocortex that sends a ``feedforward'' connection to another area, e.g., to convey sensory information, receives a reciprocal feedback connection from the same area. Furthermore, results from cognitive neuroscience indicate the same areas active during perception are also mobilized during visual imagery \cite{Kosslyn1995,Pearson2019}, suggesting a neural substrate for an internal model of the world. Other results from neuroscience point to anticipatory or predictive neural activity in various cortical areas \cite{Duhamel-1992,Umeno-Goldberg1997,Nakamura-Colby2002, fagioli2007intentional, Schneider2022}. 

\begin{figure}
\centering
\includegraphics[width=0.8\linewidth]{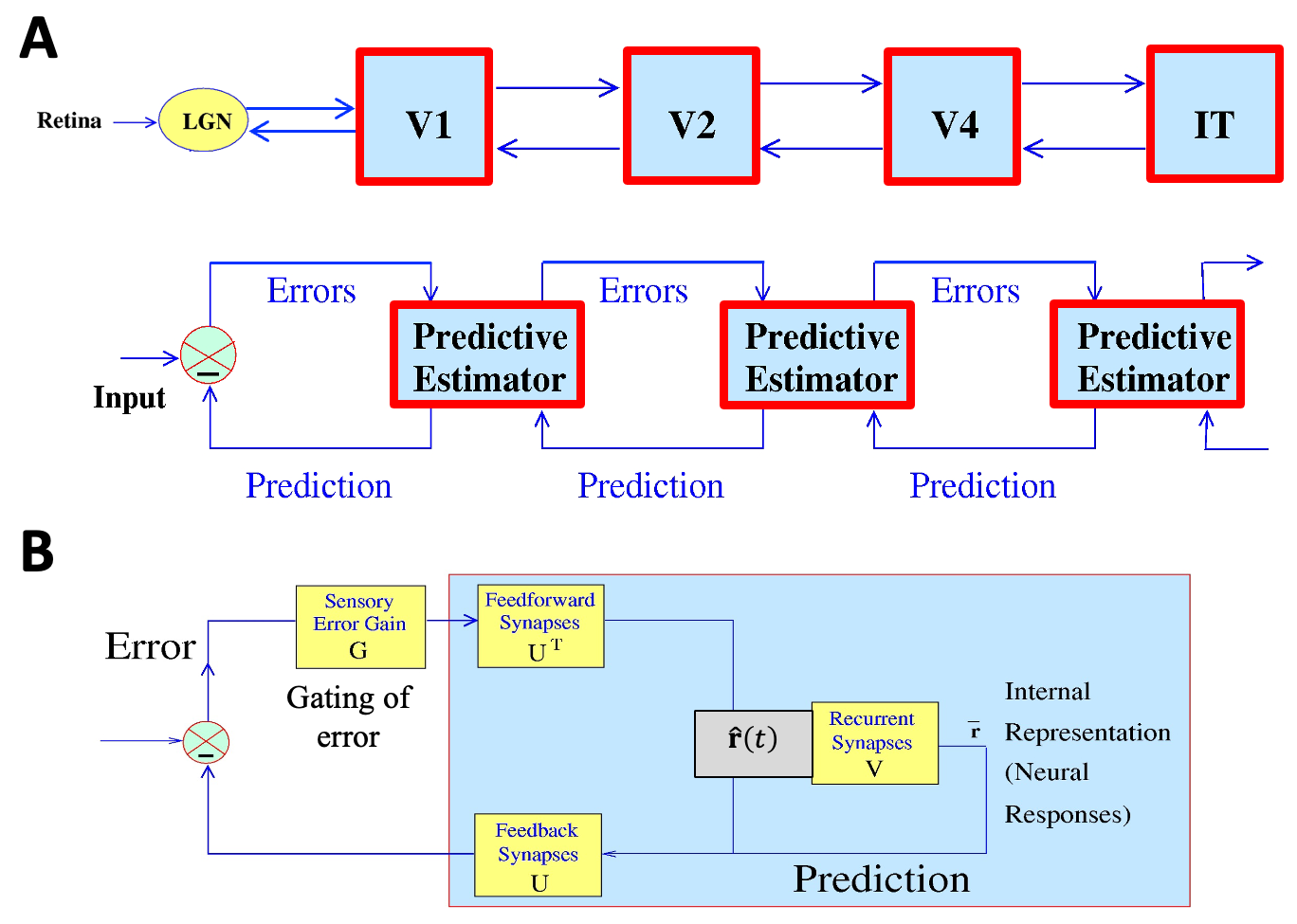}
\caption{
\textbf{Hierarchical Predictive Coding}. (A)  Hierarchical structure of a predictive coding network (bottom) emulating the hierarchy of visual areas in the cortex (top), with feedback connections carrying predictions of lower-level activity and feedforward connections conveying the errors. (B) Each ``Predictive Estimator'' module includes feedforward, feedback and locally recurrent synaptic weights that are learned using local prediction errors. These errors are also used during inference to correct the state estimates $\hat{\bf{r}}(t)$ and generate better predictions. (Adapted from \cite{Rao_Ballard_1999})
}
\label{fig:pred-coding}
\end{figure}

These results led Rao and Ballard \cite{Rao_Ballard_1999,Rao_Ballard_1997, Rao_1999} to propose a hierarchical predictive coding model of the cortex, with feedback connections between cortical areas conveying predictions of expected neural activity from higher to lower levels and feedforward connections conveying the prediction errors (Figure~\ref{fig:pred-coding}A). The model postulates that these prediction errors are used for two purposes: (1) Inference: the errors are used to quickly update the current estimate of the state of the external world maintained by the cortical area at its specific level of abstraction, allowing the cortical area to generate a better prediction at the next time step (Figure~\ref{fig:pred-coding}B); and (2) Learning: The same errors are used to gradually update the synaptic weights (with a smaller update/learning rate than inference) in the feedback and feedforward pathways as well as the local recurrent connections within the cortical area, thereby updating the internal world model. The availability of prediction errors at every level of the hierarchical network makes the learning rule biologically plausible (unlike backpropagation) while at the same time approximating backpropagation learning (it has been shown that the synaptic weight changes prescribed by predictive coding converge to those of backpropagation under specific assumptions \cite{Whittington_Bogacz_2017}).  

In predictive coding, the neural activities at each level of a hierarchical network represent the beliefs about the hidden causes responsible for input stimuli -- these beliefs are jointly influenced by both the top-down predictions from higher levels and the bottom-up error signals. Viewed within a Bayesian framework, the top-down predictions convey prior beliefs based on learned expectations stored hierarchically within the synaptic weights across the network, while the bottom-up prediction errors carry evidence from the current input. Predictive coding combines these two sources of information, weighted according to their reliability (inverse variances or “precisions”), to compute the posterior beliefs over hidden causes at each level. The objective of minimizing prediction errors across all levels can thus be shown to be equivalent to finding the maximum a posteriori (MAP) estimates of the hidden causes \cite{Rao_Ballard_1997, Rao_1999}. 

Predictive coding and the principle of prediction error minimization are closely related to variational inference and learning, which form the basis for variational autoencoders (VAEs) in machine learning research \cite{Dayan_Hinton_Neal_Zemel_1995, Kingma_Welling_2014} as well as the free energy principle (FEP) in neuroscience as proposed by Friston and colleagues \cite{Friston_2005, friston2006free, friston2009fep, friston2010fep}. Variational inference aims to find the full posterior distribution instead of a point estimate of the state as inferred via MAP estimation in traditional predictive coding. To ensure tractability, variational inference approximates the true posterior with a more tractable distribution and quantifies the “error” between the two distributions using the Kullback-Leibler (KL) divergence. It can be shown that the KL divergence reduces to the sum of the data log likelihood and a term known as the ``variational free energy''. Minimizing variational free energy with respect to the hidden states and parameters can be shown to be equivalent to minimizing the KL divergence between the approximating and true posterior distributions. This leads to the free energy principle (FEP) of brain function \cite{Friston_2005, friston2010fep, Bogacz_2017}: Within the predictive coding framework, minimizing variational free energy is equivalent to minimizing prediction errors while also attempting to be close to a prior distribution for the hidden states \cite{Jiang_Rao_2022}. Over the past decade or so, FEP and active inference (which includes actions) \cite{smith2022step}, implemented via predictive coding, has been shown to explain a wide variety of neural and cognitive phenomena \cite{Adams_Perrinet_Friston_2012, Parr_Friston_2017, Friston_DaCosta_Hafner_Hesp_Parr_2020}.

More recent work has extended the original predictive coding model in several ways. Dynamic predictive coding \cite{jiang2024dynamic} shows how the cortical hierarchy can encode temporal sequences using representations that store information over increasingly longer time scales as one ascends the hierarchy. This allows the model to explain cortical space-time response properties, postdiction in perception, and episodic memory and recall. Active predictive coding (APC) \cite{rao2024sensory,rao2023active} is a sensory-motor theory of the cortex, providing a unifying computational principle for perception, action and cognition that is inspired by the observation that the architecture of the cortex is surprisingly uniform across cortical areas. Implementations of the APC theory have demonstrated the flexibility of the model and its ability for compositional problem solving, e.g., in active vision, learning part-whole decompositions of objects, and navigation using hierarchical planning \cite{rao2023active}.

There is growing experimental evidence for predictive coding: researchers have found neurons in the visual and auditory cortex that encode predictions or prediction errors \cite{Schneider_Sundararajan_Mooney_2018, Jordan_Keller_2020,  Keller_Bonhoeffer_Hübener_2012, Fiser_Mahringer_Oyibo_Petersen_Leinweber_Keller_2016, Audette_Zhou_Chioma_Schneider_2022}. Prediction and prediction error-like signals have also been found in cortical areas in the human visual cortex \cite{Murray_Kersten_Olshausen_Schrater_Woods_2002} and the hierarchical face processing region of the monkey inferior temporal cortex (IT) \cite{Schwiedrzik_Freiwald_2017}. Other research \cite{Issa_Cadieu_DiCarlo_2018} showed that images that produced large responses early in higher-level face-selective areas in inferotemporal cortex are followed by reduced activities in lower-level areas, consistent with top-down predictions signal subduing lower-level responses as expected from predictive coding. Finally, there is also experimental evidence that cortical representations exhibit a hierarchy of timescales from lower-order to higher-order areas across both sensory and cognitive regions \cite{Murray_etal_2014, Runyan_Piasini_Panzeri_Harvey_2017, Siegle_etal_2021}.

Like predictive coding, foundation models too are trained on the objective of predicting the next input token, given past input tokens. Foundation models too are built using artificial neural networks, but rather than using biologically-plausible recurrent networks to store information about the past history of inputs, these models rely on an autoregressive approach using transformer networks: They assume all or a large fraction of past input tokens are available for use (in the form of a context window) with a self-attention mechanism (at multiple levels) for inferring embeddings that help predict the next input. They avoid the pitfalls of backpropagation through time and leverage current GPU-based compute architectures, but their predictive accuracy depends on keeping and processing huge numbers of past input tokens and training on vast amounts of data. Some of the issues with such an approach include the lack of easy interpretability of the network's behavior, lack of grounding of the concepts learned, and lack of mechanisms for control and action, as discussed in more detail below.

\section{Actions, Grounding and World Modeling} \label{sec:actions}
\begin{figure}
\centering
\includegraphics[width=0.8\linewidth]{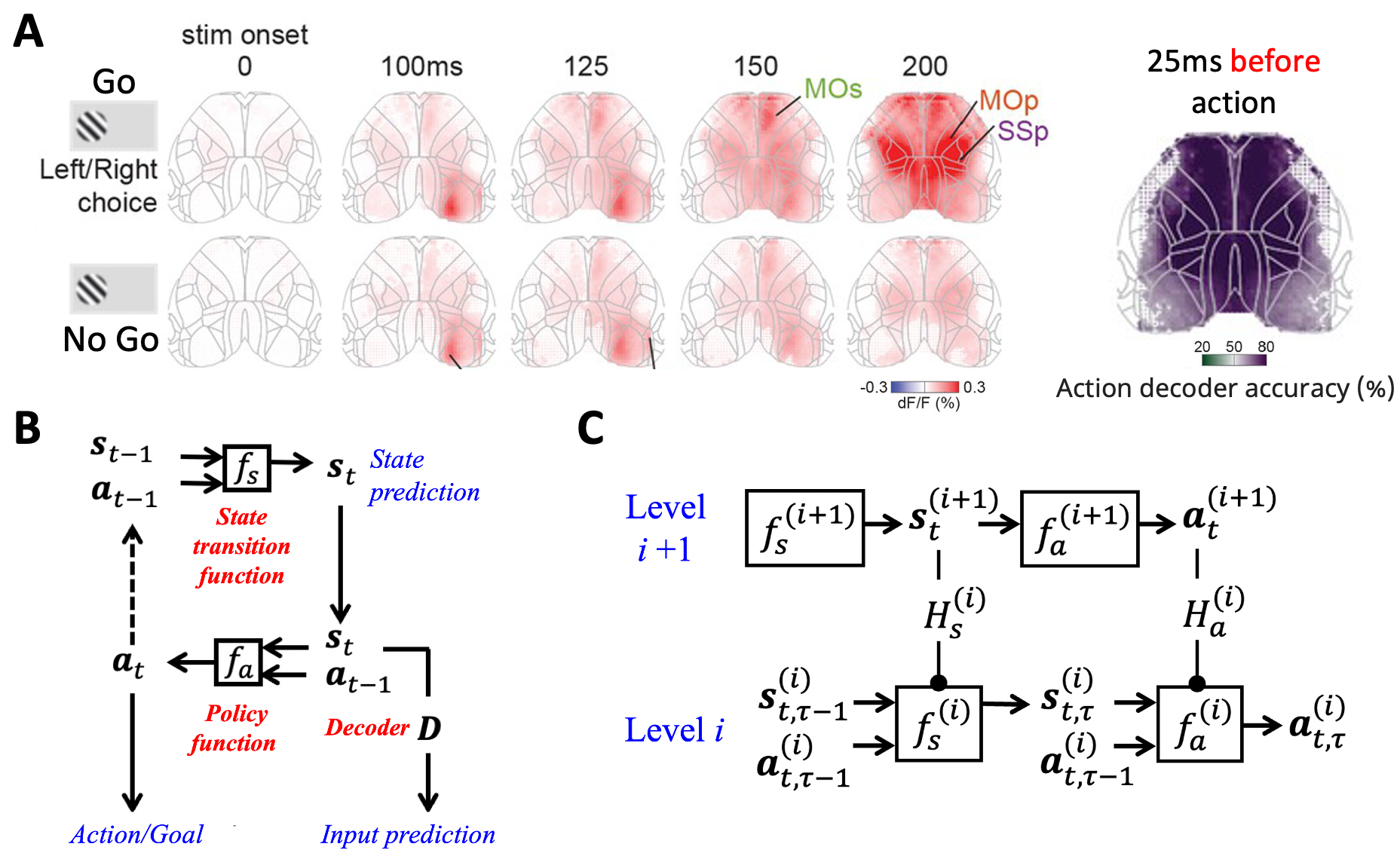}
\caption{
\textbf{Importance of Actions in the Brain and Active Predictive Coding}. (A) Almost all areas in the brain, including areas traditionally labeled as “sensory cortex,” are influenced by upcoming actions. The cross sections of the mouse brain on the left (upper row) show neural  activation across the entire cortex (measured via widefield calcium imaging) from 0 to 200ms after the onset of a stimulus and before movement in a Left/Right action selection trial (``Go''). In contrast, ``NoGo'' trials (animal does not perform movement) do not show such activity (lower row). (Right) Average action execution decoder accuracy 25ms prior to movement onset (darker pixels denote higher accuracy). The plot shows that an impending movement can be decoded from cortical activity in most imaged regions. (Adapted from \cite{Zatka-Haas2021}) (B) Active predictive coding model for a single level. (C) Hierarchical active predictive coding model. See text for details. (From \cite{rao2024sensory,rao2023active})
}
\label{fig:actions-APC}
\end{figure}

{\bf Intelligence may not have evolved without the ability to move}. Early nervous systems developed intelligence to enable an organism to sense the environment and move towards nutrients required for survival and away from noxious stimuli and predators. Several hundred million years later, we still find a tight connection between sensation and action maintained in the brains of mammals like mice and humans. For example, recent large-scale recordings across the neocortex in mice have revealed \cite{Zatka-Haas2021} that almost all areas, including areas traditionally labeled as ``sensory cortex,'' are influenced by upcoming actions (Figure~\ref{fig:actions-APC}A). Regions of the brain responsible for executing actions send ``efference copies'' of impending actions to other brain regions such as the cortex to allow them to update their neural representations in anticipation of expected sensory stimuli. 

Active predictive coding (APC) \cite{rao2024sensory,rao2023active} is a recently introduced extension of predictive coding that acknowledges the central role of actions (e.g., eye, head, body movements) in perception, cognition and behavior. The model is motivated by the fact that each cortical area, even primary sensory areas such as V1 (primary visual cortex), not only receives sensory information but also sends outputs to the brain's evolutionarily-older motor centers such as the superior colliculus and the spinal cord. The APC model assumes that each cortical area computes a “state-prediction” function (or ``world model'' \cite{ha2018world,Lecun-path-2022}) that predicts the next sensory state given a current state and action. The same area also computes an action or “policy” function that maps the current state to an action appropriate for the current task. The recurrent neuronal networks implementing state prediction and policy functions feed their outputs to each other as shown in Figure~\ref{fig:actions-APC}B, leading naturally to a sequence of predictions of states and actions. Such a sequence can track environmental states during behavior or alternatively, be used for internal simulation and planning. Sensory feedback and efference copies from the sensory organs and the motor centers respectively are sent back to update the states of higher-level networks in the neocortex. The APC model provides explanations for both the neuroanatomical structure of the neocortex as well as properties of neurons that have been observed across cortical areas (see \cite{rao2024sensory}).

The above observations from neuroscience and APC theory point to a potentially significant missing component of current transformer-based foundation models: there is no separate controller or policy network that computes actions and sends efference copies to the predictive component of the architecture to guide and ground the predictions of inputs. Consider, for example, an LLM tasked with going to the grocery store (Figure~\ref{fig:compo-examples}A) and getting avocados that are soft (i.e., ripe). The LLM can generate the sentence ``This avocado is soft'' but without the ability to act (by pressing gently) and learning the sensations associated with that action (as a human would), the LLM is left with only a superficial understanding of what it means for something to be soft (based merely on word associations) \cite{Pezzulo2024meaning,bender2021dangers}. \\
\\
\noindent \textbf{``Actions'' in current foundation models}: Modern LLMs do not use actions in the formal sense defined in reinforcement learning or neuroscience. The notion of pseudo-actions can however be introduced, either via special control tokens, as in Agentic AI \cite{yao2023react}, or by treating each generated token as an action, enabling policy-gradient-based finetuning \cite{ppo_2017_schulman, shao2024deepseekmathpushinglimitsmathematical}. One could also argue that today’s foundation models are already trained using agent-like mechanisms: Methods such as ReAct \cite{yao2023react} explicitly integrate action tokens into the model’s reasoning traces, and LLM agents leverage in-context learning to make interactive decisions without explicit policy networks \cite{yao2023react, wang2024survey, xi2025rise, patil2024gorilla}. A parallel line of work employs reinforcement learning to improve alignment and safety. RLHF \cite{christiano2017deep, ouyang2022training} shapes LLM behavior toward desirable human preferences, while RL-based finetuning of reasoning traces has been used to elicit Chain-of-Thought (CoT) reasoning \cite{Wei_Wang_Schuurmans_Bosma_Ichter_Xia_Chi_Le_Zhou_2022, wei2022emergent, kojima2022large}. In these approaches, model-free RL algorithms - most commonly Proximal Policy Optimization (PPO) \cite{ppo_2017_schulman, shao2024deepseekmathpushinglimitsmathematical} - reward correct intermediate reasoning steps, reducing hallucinations at inference time and improving performance on mathematics and program synthesis.

Despite the introduction of such pseudo-actions into foundation models, fundamental limitations remain. First, modifying the generative model itself via RLHF or other techniques means that the model is no longer faithful to its training data distribution, deviating from the goal of learning a generative model of the data. Additionally, by making a generative model for inputs generate action tokens mixes up  sensory states and actions, unlike the APC model where such a separation was shown to facilitate transfer of the generative model across tasks \cite{rao2023active}. 
CoT-based reasoning agents generalize poorly on few-shot, out-of-distribution tasks such as ARC-AGI-2 \cite{chollet2025arcagi2newchallengefrontier}, which require inferring latent generative rules from examples. They also fail on structurally simple logical and puzzle tasks (e.g., Tower of Hanoi) that humans solve effortlessly, revealing brittle task abstractions \cite{nezhurina2025alicewonderlandsimpletasks, shojaee2025illusionthinkingunderstandingstrengths, jin2025reasoninghurtinductiveabilities}. Uncertainty is miscalibrated, and hallucinations remain pervasive: LLMs confidently generate false content and often fail at self-error detection \cite{farquhar2024detecting, manakul2023selfcheckgptzeroresourceblackboxhallucination, Dahl_2024}. Finally, due to their lack of a grounded world model, these systems frequently collapse into reward-hacking or sycophantic behaviors during RL finetuning \cite{mahowald2024dissociatinglanguagethoughtlarge, niu2024largelanguagemodelscognitive, sharma2025understandingsycophancylanguagemodels, Lecun-path-2022}.\\

\noindent \textbf{Recommendations from neuroscience}: The ability to act on the world allows humans to learn causal structure, ground internal representations in sensation and motor control, and learn the physics of the world \cite{Pezzulo2024meaning,smith2005development,Buzsaki-book-2019}. This grounding enables the brain to reject internally generated predictions that violate basic physics and recognize them as hallucinations. Current LLMs, trained purely from text, lack these grounding signals. Evaluations of physical and causal reasoning in LLMs show frequent violations of simple invariants such as gravity and containment \cite{wang-etal-2023-newton,shojaee2025illusionthinkingunderstandingstrengths}. Efforts to retrofit ``world models'' into LLMs, for example by using executable code-based structures, rely on assumptions such as the availability of a description of the environment or a ``language manual'' \cite{zhang2024languageguidedworldmodelsmodelbased,dainese2024generatingcodeworldmodels}, which are not easily available for real-world environments.

\begin{wrapfigure}{l}{0.5\textwidth}
    \centering
    \includegraphics[width=0.5\textwidth]{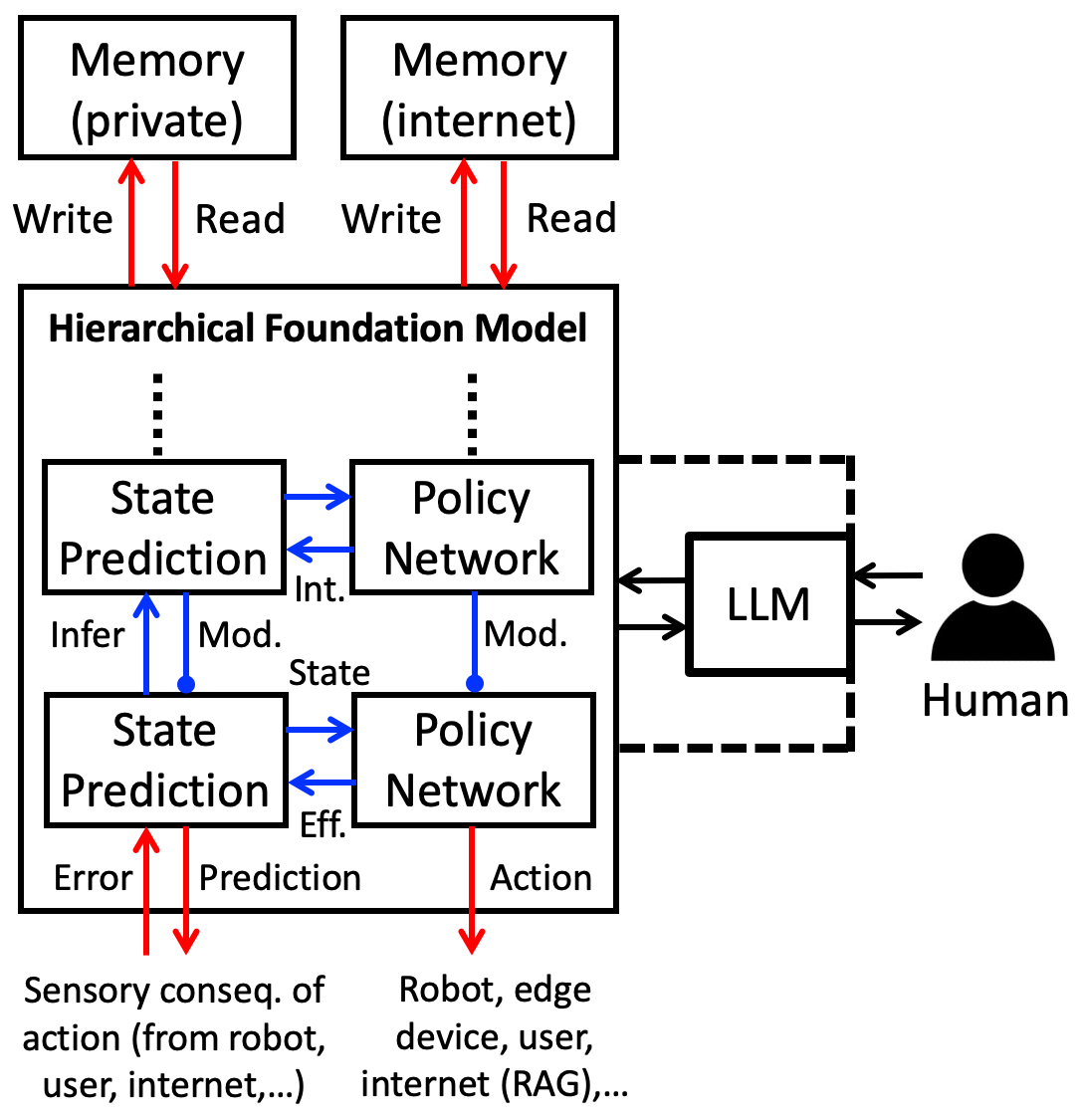}
    \caption{\textbf{A New Brain-Inspired Architecture for Foundation Models.} Mod.: modulation of lower network. Eff./Int.: efference copy/internal action for grounding/prediction control. See text for details.}
    \label{fig:summary}
\end{wrapfigure}

To address these issues, we propose restructuring or augmenting LLMs using a modular, brain-inspired architecture that combines a hierarchical world model with explicit policy networks, as shown in Figure~\ref{fig:summary}. Following the APC framework \cite{rao2024sensory,rao2023active} (Figure~\ref{fig:actions-APC}C), the world model consists of hierarchical state prediction networks, each of which could be a transformer. These networks are separated from their respective policy networks that select both external actions (for tool use or environment interactions) and internal actions (for memory queries or simulation steps). The dedicated world model maintains a multi-modal sensory latent state and learns transition dynamics under the actions prescribed by the policy networks, predicting expected sensory or internal consequences (see Figure~\ref{fig:summary}). Efference copy-like connections allow the system to compare predicted and actual outcomes, enabling grounded learning and detection of implausible inferences. 

There are two ways in which LLMs could benefit from the brain-inspired architecture described above. First, the LLM itself could be one of several vertical hierarchical state prediction networks depicted in Figure~\ref{fig:summary}, with its own policy networks controlling token generation. This hierarchical LLM would be conditioned on the latent states of other state prediction networks, e.g., vision, tactile sensation, proprioception etc. This option is indicated by the dashed box in Figure~\ref{fig:summary} indicating that the LLM is part of the hierarchical architecture. Alternately, an external third-party LLM could be augmented with a trained hierarchical world model, policy networks, and memory as shown in Figure~\ref{fig:summary}. This external LLM would consume compact summaries of the world model's multi-modal sensory latent state and in turn generate interpretable language, tool calls, and high level reasoning conditioned on the world model's latent state. Control and prediction in the physical world are solved entirely using the world model and policy networks, with the external LLM providing a flexible encoder-decoder interface for human interaction and task specification. This alternate option offers a way of leveraging the current investments in LLMs while equipping them with mechanisms for acting, grounding concepts, and acquiring stable world dynamics within a human brain-inspired architecture. Early evidence shows that world model augmented LLMs produce more coherent physical reasoning and grounded concept learning than standard LLMs \cite{zhang2024languageguidedworldmodelsmodelbased,dainese2024generatingcodeworldmodels,deepmind2025genie3}.

\section{Compositional Structure and Hierarchy of States and Actions} \label{sec:hierarchy}
Another important feature of the human nervous system is that it exhibits hierarchical organization in both sensory and motor domains (see Figures~\ref{fig:hier-brain}A and \ref{fig:hier-brain}B): this makes sense from an evolutionary point of view because evolution builds new capabilities on top of what is already available, and solutions to new problems in new ecological niches could be potentially arrived at by adding a higher level network that combines previously evolved lower-level solutions \cite{kaas2008evolution, mengistu2016evolutionary}. For example, a higher motor region in the brain can execute an ``abstract action'' that invokes a sequence of motor pattern generators in the spinal cord for walking, turning, etc. to navigate to a food source. 

\begin{figure}[t]
    \centering
    \includegraphics[width=0.8\textwidth]{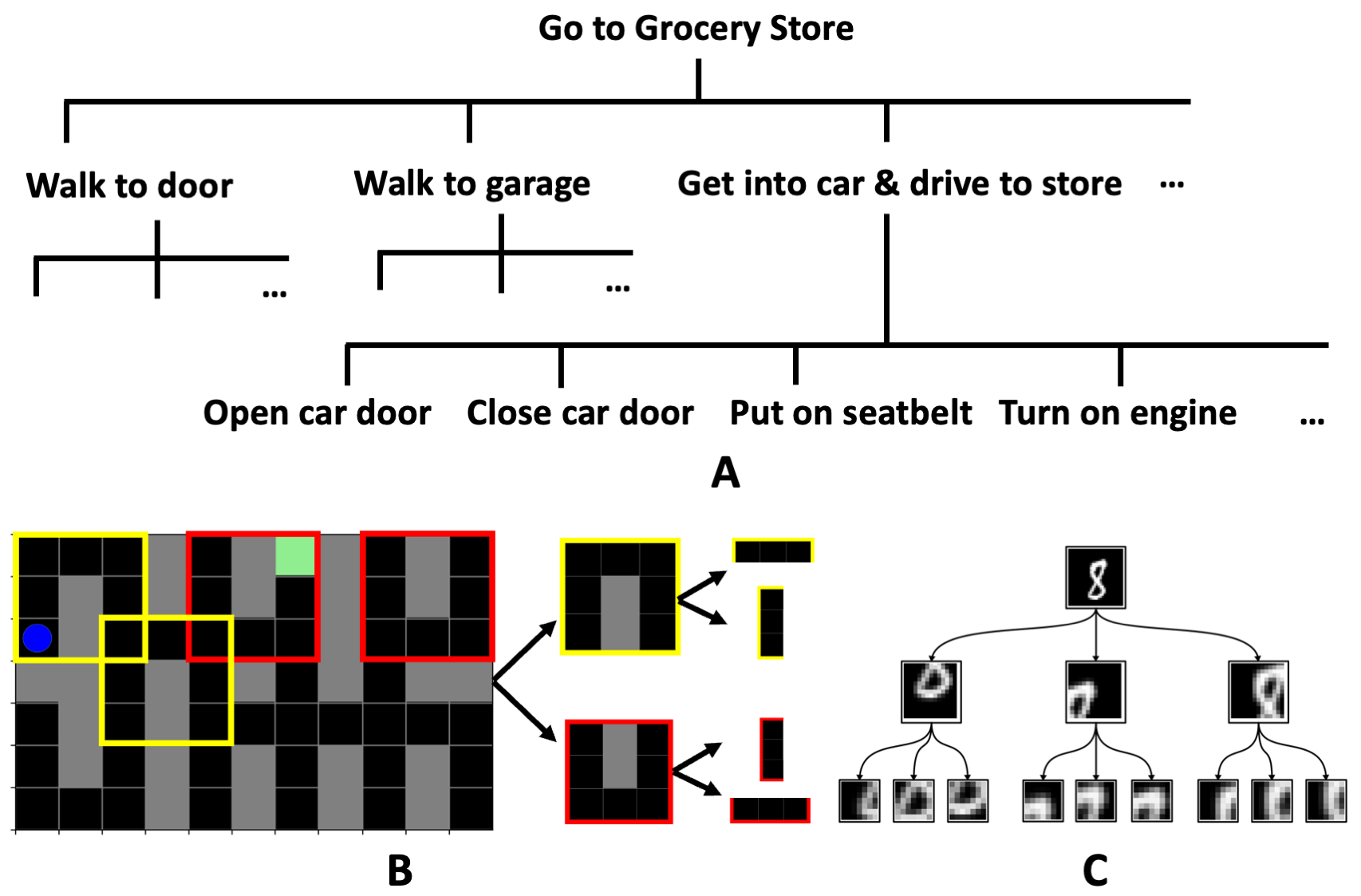}
    \caption{
    \textbf{Using Compositionality to Tackle Complex Worlds, Tasks and Scenes.} (A) Decomposition of the ``Go to Grocery Store'' problem into sub-goals/tasks, each of which can be further divided into sub-sub-goals/tasks. Note that the rate of change is faster at the lower levels compared to the higher levels, leading naturally to a temporal hierarchy. (B) A navigation problem in a maze-like building environment with corridors (black) and walls (gray), the blue dot indicating current location and green square the desired goal location. The structure of the environment can be understood in terms of its state transition dynamics, which in turn can be divided into the simpler transition dynamics of its compositional elements, two rooms outlined in yellow and red that appear at several different locations within the reference frame of the environment. These simpler elements can be further decomposed into horizontal and vertical corridors shown on the right that appear at different locations within the local reference frame of each room. (C) An object (such as a handwritten digit ``8'') can be divided into parts (loops and curves at the middle level), each of which can be divided into sub-parts (strokes, lines, smaller curves at the lower level). Each part/sub-part is associated with its coordinates (location/transformation) within a local reference frame. (From \cite{rao2024sensory})
    }
    \label{fig:compo-examples}
\end{figure}

Solving a complex problem by breaking it down into simpler sub-problems is a computationally sensible strategy as well: for example, consider again the problem (introduced in the previous section) of going to the grocery store to get avocados (Figure~\ref{fig:compo-examples}A). At a high level of abstraction, you may divide the task into sub-tasks (or ``sub-goals'') such as walking to the door of your house from whichever room you are currently in, opening the door and walking to your garage where your car is parked, getting into car and driving to the grocery store, parking your car, walking to the entrance of grocery store, etc. Note that many of these sub-tasks can be {\em re-used} for solving a range of other problems (e.g., going to work, going to visit a friend, going to a restaurant, etc.). Therefore, it may be useful to learn and maintain a policy for the sub-task to avoid planning actions each time the sub-task is re-used as part of a different task; indeed, this is precisely the motivation behind well-known hierarchical reinforcement learning (HRL) frameworks such as options \cite{sutton1999between, barto2003recent, abel2022theory}. The ability to divide a problem into components and re-use the components to solve new problems gets to the heart of {\em compositionality}, a powerful attribute of representations that may form the basis for the flexibility and fast generalization ability of human cognition \cite{lake2017building,smolensky2022neurocompositional, ellis2021dreamcoder, andreas2019measuring}. 

Note that compositionality for an agent interacting with the physical world can involve not only dividing a complex action into simpler actions but also dividing a complex state, defined by its state transition function, into simpler state transition functions. This is illustrated in the example in Figure~\ref{fig:compo-examples}B, a simple version of the ``Go to grocery store'' problem: the problem is to navigate in a maze-like building environment with corridors (black) and walls (gray) (blue indicates the current location and green the current goal location). This environment can be divided into its compositional elements, two simpler rooms outlined in yellow and red that appear at different locations as shown in Figure~\ref{fig:compo-examples}B. These simpler rooms can in turn be decomposed into horizontal and vertical corridors as shown in the figure. Note that the rooms are ``simpler'' in the sense that they have simpler state transition functions than the original building environment, and the corridors in turn have simpler state transition functions than the rooms.

Interestingly, the same concept of decomposing a task into sub-tasks, sub-sub-tasks, etc.\ as described above can be applied to the visual perception problem of representing an object in terms of parts, sub-parts etc. For example, a person can be represented in terms of the parts: head, arms, torso, and legs; a leg in turn is made up of sub-parts such as the foot, ankle, lower leg, knee, upper leg etc. Similarly, a handwritten digit, e.g., an ``8'', can be divided into parts (loops and curves), each of which can be divided into sub-parts (strokes, lines, smaller curves) as shown in Figure~\ref{fig:compo-examples}C. 
 
For the APC model described in the previous section, hierarchical states and actions can be obtained as follows (Figure~\ref{fig:actions-APC}C): a state representation {\em vector} $s^{(i+1)}$ at abstraction level $i+1$ generates or modulates (via a hypernet $H_s^{i}$ \cite{Hypernetworks-ref-2017,rao2024sensory}) a state transition {\em function} $f_s^{(i)}$ at the lower level $i$ (along with an initial state vector $s_0^{(i)}$ to start the lower-level state-action sequence); similarly, a higher-level action {\em vector} $a^{(i+1)}$ generates (via a hypernet $H_a^{i}$) a lower-level policy {\em function} $f_a^{(i)}$ (``option'' in RL \cite{sutton1999between, abel2022theory}). The two lower-level functions interact with each other, as in Figure~\ref{fig:actions-APC}B, to generate lower-level states and actions as shown in Figure~\ref{fig:actions-APC}C. Each such state vector and action vector can in turn generate transition and policy functions at an even lower level of abstraction. A lower-level sequence executes for a period of time until a condition is met (e.g., a sub-goal is reached, a task is completed or times out, or there is an irreconcilable error at that level). Then, control returns to the higher-level, which transitions to a new higher-level state and a new higher-level action, and the process continues. It is clear that such a generative model can generate the dynamics of the states (the ``physics'' of the world) and action sequences (produced by a policy) at different time scales, providing a mathematical and hierarchical way of describing complex tasks such as going to the grocery store.
\\\\
\noindent \textbf{Compositionality and hierarchy in existing foundation models}: Explicit attempts to introduce hierarchical structure into transformers have been explored, motivated largely by the practical need to reduce the very large number of input tokens (``context window'') needed in the autoregressive architecture of transformers to obtain good prediction performance. Most current transformer-based foundation models rely on a moving context window that keeps the number of input tokens tractable \cite{beltagy2020longformer, zhang2019hibert, liu2021swin, chalkidis2022exploration, correia2023hierarchical}. It can be argued that transformers already possess a multi-layered architecture through the stacking of self-attention blocks, which can allow compositional structure to emerge during training on large-scale internet data that is compositional in nature. Prior work has investigated this intrinsic compositionality by projecting transformers into tree-structured representational spaces, finding that their behavior becomes increasingly tree-like over the course of training \cite{Murty2023,schug2025attention,McCoy2020,garg2022can}. Similarly, advances in CoT prompting and multi-step reasoning have improved the ability of LLMs to decompose complex problems into manageable subproblems \cite{OpenAI_2024, Wei_Wang_Schuurmans_Bosma_Ichter_Xia_Chi_Le_Zhou_2022, yao2023react}.

Nevertheless, these trends leave several limitations unresolved. The long input context window is still processed as a single flat buffer rather than as structured episodes, leading to fragile handling of past information. Specifically, such models show a strong “lost in the middle” failure mode: accuracy drops sharply when relevant information is positioned in the middle of the context window, revealing weak hierarchical organization and poor content-based retrieval \cite{liu-etal-2024-lost}. Existing hierarchical transformers typically downsample or pool for computational efficiency rather than inducing reusable abstractions as in APC. Planning benchmarks such as PlanBench demonstrate that even with full access to domain descriptions and action histories, LLMs perform poorly on long-horizon tasks \cite{valmeekam2023planbenchextensiblebenchmarkevaluating}. CoT prompting also fails to address these structural problems. Evaluations on PlanBench and related datasets show that LLMs systematically struggle with multi-step planning, regression, and reasoning about actions and state transitions, even in domains where symbolic planners succeed with ease \cite{valmeekam2023planbenchextensiblebenchmarkevaluating, lanham2023measuringfaithfulnesschainofthoughtreasoning, wang2025hierarchicalreasoningmodel}.
\\\\
\noindent \textbf{Recommendations from neuroscience}: Transformer-based foundation models remain monolithic and are not explicitly organized to support hierarchical or compositional processing in a manner comparable to biological systems or to the structure of the APC model. In contrast, hierarchical predictive coding likely emerged in biological systems as an energy-efficient solution to neural computation \cite{bakhtiari2022energyefficiency,ali2022energyefficiency, friston2009fep}. Motivated by this, we advocate replacing the flat context buffer used in current transformer models with multi-timescale context processing. A hierarchical buffer should maintain distinct fast and slow contextual representations, detect event boundaries, and segment token streams into compositional units suitable for compressed processing \cite{HASSON2015304, NEURIPS2023_streamer}. The APC architecture described above naturally extends to hierarchical representations of states and actions within the latent spaces of world models and policies. Hierarchical RL and the options framework offer a concrete route for learning rich hierarchical policies \cite{BOTVINICK2012956, sutton1999between, barto2003recent}. Integrating current LLMs with these APC components at each level of hierarchy would enable them to encode and decode structured interpretable representations, while substantially improving their processing and planning capabilities.

While CoT prompting does enable superficial task decomposition, we hypothesize that foundation models equipped with hierarchical predictive coding-like mechanisms will be significantly more energy efficient and more robust to compositional generalization than today’s monolithic architectures \cite{OpenAI_2024, Wei_Wang_Schuurmans_Bosma_Ichter_Xia_Chi_Le_Zhou_2022, yao2023react, crownhart2024energyuse, Lecun-path-2022}. Hierarchical routing modules such as those used in GPT-5 \cite{openai2025gpt5} functionally behave in a manner similar to the proposed APC components. These modules route inputs to appropriate expert models based on context and anticipated compute expenditure, and learn via feedback. We believe that such modifications are consistent with our recommendations and represent a promising direction for instantiating hierarchical processing in foundation models.

\section{Episodic Memory} \label{sec:memory}
If the cortex consists of a hierarchy of state and action networks, what is at the top of this hierarchy? Anatomically, as shown in Figure~\ref{fig:hier-brain}A (left), the hierarchy terminates with the hippocampus, an evolutionarily ancient brain structure that has long been associated with episodic memory: such memories can range from a recent rewarding navigational memory (in a rodent) to an unexpected interaction with a stranger earlier in the day (in a human). The hippocampus is thought to store memories of salient events that occur during the day, consolidating those memories to the cortex via learning during rest and sleep \cite{Mcclelland1995CLS}. It is also known that prediction errors (or “surprise”-related signals), as used in predictive coding models, can drive memory reactivation and reconsolidation \cite{Kim_Lewis-Peacock_Norman_Turk-Browne_2014, Rust_Palmer_2021}.

One advantage of having a memory for storing recent experiences (episodes) at the top of a predictive coding hierarchy is that such a memory can provide a longer context for prediction than can be stored in the hidden states of recurrent networks in the hierarchy. Storing such context is critical for solving complex tasks such as our ``Go to Grocery Store'' problem (Figure~\ref{fig:compo-examples}A) where the current higher-level goal and sub-goals must be maintained while predicting states and executing actions at a lower level. Unlike LLMs and other foundation models, the cortex cannot keep a vast number of past input tokens for predicting future inputs, but uses the parsimonious strategy of using a memory (hippocampus) to maintain historical context for effective prediction in a hierarchy of recurrent networks (cortex). A second advantage is that storing salient episodes in a memory allows replay of the episodes, or more usefully, playback of novel combinations of memory snippets \cite{Gupta2010,Pfeiffer2013} for training the cortical network when the animal is at rest. This playback is akin to data augmentation, exploiting the compositional nature of the world and allowing the cortical network to learn to generalize to novel scenarios and new scene/event compositions not encountered (yet) in the real world \cite{Kurth-Nelson2022}. 

Previous work in both theoretical neuroscience and AI has leveraged the idea of using an external memory to augment recurrent networks \cite{Santoro_Bartunov_Botvinick_Wierstra_Lillicrap_2016}. The dynamic predictive coding model of cortex \cite{jiang2024dynamic} (described above) and the Tolman-Eichenbaum machine model of cortex-hippocampus interactions \cite{Whittington_Muller_Mark_Chen_Barry_Burgess_Behrens_2020} are examples of neuroscience models using Hopfield-style autoassociative memories for storing current episodic context and predicting inputs (see also \cite{Raju_Guntupalli_Zhou_Wendelken_Lázaro-Gredilla_George_2024}). In AI, neural network architectures such as Neural Turing Machines \cite{Graves_Wayne_Danihelka_2014} and MERLIN \cite{Wayne_etal_2018} have demonstrated that allowing recurrent networks to learn to use an external memory significantly increases their computational capabilities. 
\\\\
\noindent \textbf{Episodic memory in existing foundation models}: Foundation models based on transformers play a role similar to the cortex in terms of learning a model of the world (e.g., from language and/or  vision) using predictions and prediction errors. However,  transformers are autoregressive and thus, unlike the cortex, an ``episodic memory'' of recent past inputs is always available to the model in the form of the current context window (state space models such as MAMBA \cite{gu2023mamba} are not autoregressive and are  closer to biological plausibility in this regard). There has also been considerable recent interest in LLMs capable of retrieval augmented generation (RAG) \cite{Lewis_etal_2020, guu2020retrieval, borgeaud2022improving, Shi_Min_Yasunaga_Seo_James_Lewis_Zettlemoyer_Yih_2024}, which allows an LLM to retrieve information from a private database or from the internet. This could supply the LLM with information beyond what is available in the context window. Parallel efforts to elicit in-context learning in robotics have given rise to Vision-Language-Action models (VLAs) that use historic episodes as inputs \cite{lin2025onetwovlaunifiedvisionlanguageactionmodel, zhao2025cotvlavisualchainofthoughtreasoning, schuurmans2023memory, sarch2023open}.

However, transformer-based foundation models only process a recent window of tokens for practical reasons, limiting their episodic memory capacity. Both long-context attention (discussed in~\ref{sec:hierarchy}) and RAG models treat past inputs as a flat query-driven buffer and as a consequence, are susceptible to the “lost in the middle” problem \cite{liu-etal-2024-lost}. RAG models, which rely on embedding-based nearest-neighbor retrieval, often retrieve topically related but misleading documents, and continue to suffer from hallucinations or confabulations \cite{huang2023hallucination, zhang2025raghallucination, magesh2025legalrag}. There is also no principled multi-timescale organization of episodic memory in current foundation models. Biological evidence suggests that memory is an intrinsic component of information processing distributed across a hierarchy of cortical timescales, with different areas integrating information over progressively longer windows \cite{HASSON2015304, baldassano2017discovering}.
\\\\
\noindent \textbf{Recommendations from neuroscience}: We argue that foundation models should be augmented with the ability to not only retrieve information from a memory (as in RAGs) but also write salient information or episodes to the memory. This would bring foundation models closer to the Neural Turing Machine and MERLIN models discussed above, but would replace the recurrent networks in these models with more powerful large-scale transformers and a mechanism (e.g., prediction-error driven) for intelligently selecting which episodes or parts of episodes to store in memory. Additionally, the AI system could emulate the sleep consolidation idea from neuroscience by utilizing these stored memories for compositional playback of episodic snippets to the foundation model during an offline ``sleep'' phase, enabling the model to continually update its knowledge and remain up-to-date using this brain-inspired form of intelligent data augmentation.

While RAGs and memory-augmented VLAs do have access to recent episodic context, this context is limited to a manageable window of recent tokens since storing all input tokens since the beginning of time (or since the beginning of an episode or interaction with a user) is impractical even for the largest models. As a result, any experience or user interaction that occurred before beginning of this window is ``forgotten'' by the foundation model. In the case of VLAs, the historical context used is short and typically not writable. We argue that these shortcomings of current models can be remedied by storing the model's experiences beyond the horizon of the current context window in a separate memory or neural network for salient episodes. This selective memory of past episodes can be used to (1) augment the traditional multi-level attentional processing of the current context window, and (2) as described above, enable novel playback and continual learning of the foundation model to keep improving its performance in the long term.

\section{Discussion and Conclusion} 
\label{sec:discussion}
Reading some of the papers by the early pioneers of AI (e.g., in \cite{shannon1956automata}) makes one realize how much these pioneers were inspired by the major discoveries being made in brain science at that time. The foundation models of today owe a debt to the mathematical abstraction of neurons proposed by McCulloch and Pitt \cite{mcculloch1943logical} as well as to Rosenblatt's perceptron framework \cite{rosenblatt1958perceptron}. Convolutional networks were inspired by neuroscientists Hubel and Wiesel's description of repeating feature-detecting modules across the visual cortex, while the attention operator used in transformers emulates ``covert attention'' in the brain. The infusion of new ideas from neuroscience into AI has diminished in recent years, being replaced by a race to develop new AI frameworks built on increasingly larger GPU-compute architectures and increasing amounts of data on the internet. We now have foundation models with billions of learned parameters and human-like natural language and reasoning capabilities, but without the ability to ground their learned concepts in the real world through actions, and without learning the real-world consequences of performing actions. The missing sense of agency/responsibility is complemented by a lack of explicit closed-loop control and interpretability, leading to worries about bias, safety and human-compatible behavior. 

In this perspective article, we argued that some of the above issues with present day foundation models can be ameliorated by augmenting foundation models with actions, compositional and hierarchical structure, and episodic memory, emulating what neuroscience is teaching us about some of the most salient features of the mammalian brain. Our proposal shares similarities with other hierarchical architectures for human-like intelligence, such as Albus' theory of intelligence \cite{Albus1991OutlineFA}, Friston's free-energy minimization \cite{friston2009fep}, Sutton's OaK architecture \cite{Sutton-Oak-2023} and LeCun's joint embedding predictive architecture \cite{Lecun-path-2022}. Our proposal adopts salient aspects of these prior perspectives, such as hierarchical representations and predictive coding, while leveraging insights from recent advances in neuroscience, spanning neuroanatomy, neurophysiology and behavior. 

In the long term, we believe that the quest for safe, efficient, and interpretable human-like AI can significantly benefit from supercharging the historically fruitful cross-fertilization of ideas between brain science and AI.




\bmhead{Acknowledgements}
This research was supported by the Air Force Office of Scientific Research (AFOSR) grant no.\ FA9550-24-1-0313, the National Science Foundation (NSF) EFRI Grant no.\ 2223495, and a “Frameworks” grant from the Templeton World Charity Foundation.

\bibliography{sn-bibliography}

\end{document}